# Gradient-based multi-focus image fusion with focus-aware saliency enhancement

Haoyu Li , XiaoSong Li*

**Abstract** Multi-focus image fusion (MFIF) aims to yield an all-focused image from multiple partially focused inputs, which is crucial in applications cover surveillance, microscopy, and computational photography. However, existing methods struggle to preserve sharp focus-defocus boundaries, often resulting in blurred transitions and focused details loss. To solve this problem, we propose a MFIF method based on significant boundary enhancement, which generates high-quality fused boundaries while effectively detecting focus information. Particularly, we propose a gradient-domain-based model that can obtain initial fusion results with complete boundaries and effectively preserve the boundary details. Additionally, we introduce Tenengrad gradient detection to extract salient features from both the source images and the initial fused image, generating the corresponding saliency maps. For boundary refinement, we develop a focus metric based on gradient and complementary information, integrating the salient features with the complementary information across images to emphasize focused regions and produce a high-quality initial decision result. Extensive experiments on four public datasets demonstrate that our method consistently outperforms 12 state-of-the-art methods in both subjective and objective evaluations. We have realized codes in https://github.com/Lihyua/GICI

## 1    Introduction

Depth of field [1] significantly influences the range of clear imaging achievable by an optical system and reflects the overall imaging capability of that system. Due to the limited depth of field of a camera, only part of the scene is in focus while the rest of the scene appears blurred, which may hinder the complete presentation of useful information in the scene. The MFIF technology effectively solves this problem. This technique can synthesize images from different sources into one image of the same scene to produce a clear fused image. Currently, MFIF techniques can be broadly categorized into deep learning-based methods and traditional methods. Among them, deep learning-based methods mainly include classification and regression models, while traditional methods can be categorized into spatial domain methods, transform domain methods, and methods combining transform and spatial domains.

* Corresponding author.
    E-mail addresses: lixiaosong@buaa.edu.cn

Liu et al. [2] introduce deep learning into MFIF by learning a convolutional neural network model (CNN) to generate fusion rules and activity level measurements, strengthening their correlation and avoiding traditional manual design issues. Subsequently, deep learning models like CNN, Generative Adversarial Network model (GAN) and Self-Attention Mechanism model have been effectively applied in the domain of MFIF, yielding impressive results. In the classification model-based methods, Xiao et al.[3] used multi-scale features and attention mechanism to achieve accuracy for boundary segmentation, and proposed a global feature coding U-Net, which can obtain global semantic and boundary information more accurately. In contrast, among the based regression models, Zhu et al. [4] throwed in a generalized model built upon expert mixtures - task-customized adapter mixing, which can be utilized across a wide range of fusion tasks by adding only 2.8% of learnable parameters.

Traditional methods still hold considerable importance in the domain of MFIF. Among these, spatial domain-based methods directly process and fuse images in the spatial domain, obtaining features from the original images to assess activity levels, which are then used in fusion rules to combine the images based on the detected activity. For instance, You et al. [5]introduced a pixel-level fusion method called LSDGF1, which operates on the principle that clear pixels typically have a higher local variance, resulting in a higher local standard deviation. Unlike spatial domain methods, transform domain methods approaches involve three steps: image transformation, coefficient fusion and inverse transformation. Burt PJ et al. [6] applied the Laplacian pyramid in the MFIF field by using a Gaussian basis function as the pattern element, ensuring efficient and accurate transformations. Although transform domain methods better preserve image details, the conversion process is time-consuming and prone to errors, often resulting in artifacts and blurring.

Lately, significant advancements in MFIF have produce high-quality fused images, yet inherent issue like blurred focus/defocus boundaries and loss of detail remain. In response, this paper proposes a MFIF method based on significant boundary enhancement. First, an initial fusion image is obtained in the gradient domain, and saliency map is generated from both the initial fusion and the source image using Tenengrad [7] gradient saliency detection. Subsequently, we introduce a focus detection scheme based on gradient information and complementary information (GICI) to enhance the focus areas., and finally compares it with the saliency map of the initial fused image to obtain the initial decision map, and obtains the final fused image through post-processing methods. Comprehensive experimental results show that our method surpasses existing techniques, generating fused images with enhanced edge information and successfully mitigating problems like boundary detail loss and edge blurring.The key contributions of this work are as follows:

- We propose a MFIF method based on significant boundary enhancement to effectively solve blurring and detail loss at both focused and unfocused edges.
- We develop a novel focus detection scheme, which accurately distinguishes piexl focus attributes through gradient and complementary information, ensuring clear details retention in the fused image.

- We introduce a high-quality initial decision maps acquisition strategy via comparing the salient image of the enhanced source images, effectively preserves the details and edges of the source images.

The paper is organized as follows : Section 2 details the proposed method; Section 3 describes the experimental setup, results and discussion; Finally, Section 4 concludes the study.

## 2 Methodology

Fig. 1. illustrates the proposed MFIF method, which is divided into two stages: (1) enhancement of the saliency map and acquisition of the initial decision map, and (2) optimization of the initial decision map and acquisition of the fused image. The following sections provide a detailed introduction and comprehensive analysis of each component of the model.

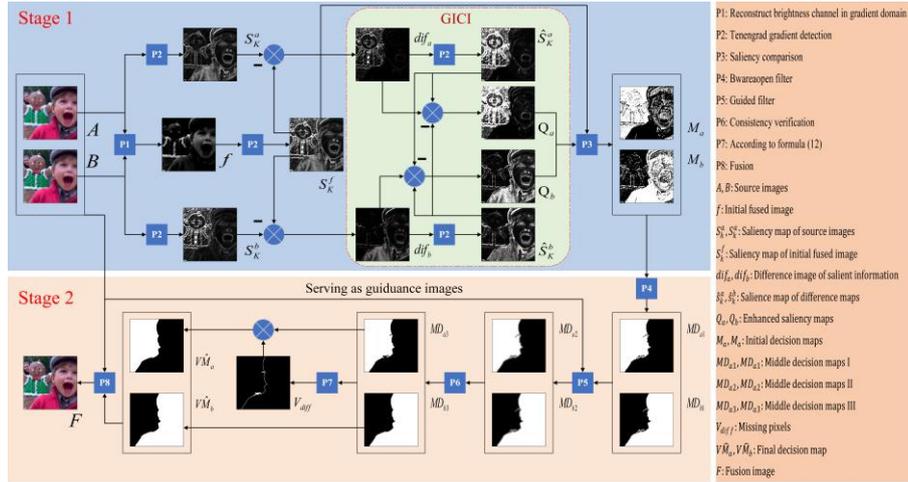

**Fig.1** Framework of the proposed MFIF algorithm

### 2.1 Acquisition of the initial decision map

To address the issue of unclear edge detection in spatial domain methods, we propose an algorithm based on saliency information enhancement, which processes the saliency map of the input image to obtain a more accurate decision map. In transform domain methods, we find that gradient-based fusion methods are particularly effective in preserving the details of the source image. Therefore, we adopt the approach proposed by Paul S. et al. [8] to obtain the initial fused image $f$, which serves as a foundation for significant information enhancement. In this method, we focus exclusively on brightness fusion, as the brightness channel contains the primary structural details of the original image. This reduces the computational burden and improves efficiency.

Typically, the gradient of the source image brightness channel can be obtained as:

$$\Phi_\alpha^x(x,y) = \alpha(x+1,y) - \alpha(x,y) \tag{1}$$

$$\Phi_\alpha^y(x,y) = \alpha(x,y+1) - \alpha(x,y) \tag{2}$$

For ease of discussion, we set the number of input images to 2 . The extension to cases with multiple input images can be easily derived based on the two-image input formulation. Let $\alpha = A$ or B denote the input source images A and B, respectively. $\Phi_n^x(x,y)$ and $\Phi_n^y(x,y)$ are the gradient components of the image in the x and y directions, while $(x,y)$ represents the pixel location in the image. Therefore, according to the method of Paul S. et al., we obtain the initial fusion image $f$.

According to the evaluation by Huang et al. [7], the Tenengrad detection method is highly sensitive to pixel variations, yielding clear and precise edge detection results. Therefore, we select the Tenengrad Gradient detection to perform significance measurements. This approach extracts gradient and edge details from the image, generating the saliency maps for input images A and B, as well as the initial fused image $f$, as shown below:

$$S_K^\beta(x,y) = \sum_{x=2}^{M-1} \sum_{y=2}^{N-1} |\nabla\theta(x,y)|^2 \quad \theta \in A, B, f \quad \beta \in a, b, f \tag{3}$$

After obtaining the saliency maps $S_K^a$, $S_K^b$, and $S_K^f$ for input images A and B and the initial composite image $f$, we calculate the differences between $S_K^f$ and $S_K^a$, and $S_K^f$ and $S_K^b$, respectively. Since the initial composite image $f$ is fully focused, this difference operation accentuates the original defocused regions in the input images, allowing them to stand out in the difference maps:

$$dif_i = S_K^f - S_K^i \quad i \in a, b \tag{4}$$

The difference images $dif_a$ and $dif_b$ are generated and used as inputs to the GICI module to differentiate focused and defocused areas in the image. The introduction of Tenengrad saliency information detection makes the features and details of the focused area in the saliency map richer, while reducing the features and details of the defocused area:

$$\hat{S}_K^i = \sum_{x=2}^{M-1} \sum_{y=2}^{N-1} |\nabla dif_i(x,y)|^2 \quad i \in a, b \tag{5}$$

We obtain the saliency maps $\hat{S}_K^a$ and $\hat{S}_K^b$ from the difference images. Compared to directly extracting saliency maps from the source images, $\hat{S}_K^a$ and $\hat{S}_K^b$ provide a clearer distinction between defocused and focused areas, along with smoother boundaries. However, small alternating regions of focus and defocus near edges can result in unclear boundaries. To address this, we propose leveraging complementary information between images by combining saliency maps $S_K^a$ and $S_K^b$ with $\hat{S}_K^a$ and $\hat{S}_K^b$ from the difference images:

$$Q_a = S_K^a + \hat{S}_K^a - \hat{S}_K^b \times k \tag{6}$$

$$Q_b = S_K^b + \hat{S}_K^b - \hat{S}_K^a \times k \tag{7}$$

Among them, $k$ serves to determine the extent of $\hat{S}_K^b$ or $\hat{S}_K^a$ to be subtracted, and $k$ is set to 0.5 through parameter analysis. Through the above operation, the saliency map of the input image is enhanced in the focus area, and the initial fusion image obtained through the gradient domain can retain more details at the boundary and reduce the introduction of error information. Therefore, we choose f and the enhanced saliency maps $Q_a$ and $Q_b$ to obtain an initial decision map that has clearer edge details and retains richer edge details：

$$M_a(x,y) = \begin{cases} 1 & \text{if } Q_a(x,y) \geq S_K^f(x,y) \\ 0 & \text{otherwise} \end{cases} \tag{8}$$

## 2.2 Optimization and acquisition of the fused image

To address potential errors that may occur during the initial decision map acquisition process, we employ the adaptive threshold "bwareaopen" filling filter to refine the initial decision map and rectify inaccurate pixels:

$$\widehat{M_a} = \text{bwareaopen}(M_a, t) \tag{9}$$

The filling filter eliminates all small areas smaller than $t$ pixels in the initial decision map. However, if the size of $t$ is set directly, the advantages of "bwareaopen" cannot be used due to the diversity of source image sizes. Therefore, to implement an adaptive threshold for $t$ and maximize the advantages of "bwareaopen", we define the threshold t as $t = th \times S$, $S$ is the size of the input image, and $th$ is selected as 0.02 through multiple parameter experiments.

To preserve edge information, we introduce Guided Filtering:

$$FM_i = GF(\eta, \widehat{M_\iota}, r, \varepsilon) \quad \eta \in A, B \tag{10}$$

where r and ε set to 5 and 0.3, respectively. To address the error information that may be introduced by Guided Filtering, we apply consistency verification to smooth and naturalize the decision map edges:

$$VM_i = \begin{cases} 1 & \text{if} \sum_{(x,y) \in \varphi} FM_i(x+a, y+b) \geq \frac{\varphi}{2} \\ 0 & \text{otherwise} \end{cases} \tag{11}$$

where $\varphi = q \times S, q = 5 \times 10^{-5}$. However, performing separate consistency verifications on the decision maps can lead to some pixels remaining unclassified in the focus-defocus boundary areas of the binary image. Therefore, we first find the pixels whose pixel value is 0 and whose focus attribute cannot be determined, and assign a value to this part:

$$V_{\text{diff}}(x,y) = \begin{cases} 1 & \text{if } VM_a(x,y) = VM_b(x,y) = 0 \\ 0 & \text{otherwise} \end{cases} \tag{12}$$

$$V\widehat{M}_a(x,y) = \begin{cases} 0 & \text{if } V_{\text{diff}}(x,y) = 1 \\ VM_a(x,y) & \text{otherwise} \end{cases} \tag{13}$$

$$V\widehat{M}_b(x,y) = \begin{cases} 1 & \text{if } V_{\text{diff}}(x,y) = 1 \\ VM_b(x,y) & \text{otherwise} \end{cases} \tag{14}$$

Then the final decision graphs $V\hat{M}_a$ and $V\hat{M}_b$ are obtained, ensuring that the addition of the two decision graphs is a binary graph with a value of all 1 and a size of $S$. This final decision map enhances the source image features, preserves edge details, and minimizes the introduction of error information.

The fused image is then generated from the final decision maps $V\hat{M}_a$ and $V\hat{M}_b$:

$$F(x, y) = V\hat{M}_a(x, y) \times A(x, y) + V\hat{M}_b(x, y) \times B(x, y) \qquad (15)$$

## 3    Experiment

### 3.1    Experimental settings

We validate our method on four publicly available MFIF datasets: the Lytro dataset [9], the MFFW dataset [10], the MFI-WHU dataset [11] and the GrayScale dataset[12]. For our experiments, we selected a representative image pair from Lytro dataset—"lytro-01"—for qualitative and quantitative analysis. Additionally, quantitative comparisons were conducted across all four public datasets.

To assess the usefulness of the proposed framework, we provide comparisons with 12 advanced methods. These methods encompass spatial domain methods: INS[13], IFD[14] ,MFIF-MMIF[15] and RDMF[16]; transform domain method: SAMF[17]; based on empty Methods that combine the inter-domain and transform domains: SIGPRO[18]; and deep learning methods: PMGI[19], MFIF-GAN[20], U2Fusion[21], MUFusion[22], SDNet[23] and MFEIF[24]. In this paper, VIF[25], $Q^{AB/F}$[26], SF[27], NMI[28], $Q_Y$[29], and $Q_{CB}$[30] metrics are selected for comparative analysis to objectively evaluate the fusion results. In the experiments, red indicates first place, green indicates second, and blue indicates third.

### 3.1    Qualitative comparisons

Due to the space limitations, we only present qualitative comparison using the Lytro dataset.

As shown in Figure 2, it is evident that methods such as PMGI, INS, MUFusion, SDNet, MFEIF, and IFD retain significant residual information in the out-of-focus areas, suggesting their inability to accurately delineate focus regions and leading to detail loss along image edges. Furthermore, the difference maps reveal that MUFusion and MFEIF produce excessively bright outputs with lingering background textures, and IFD suffers from noise that blurs pseudo-color map details. These observations highlight that these methods struggle to precisely identify focus areas and maintain accurate edge representation.

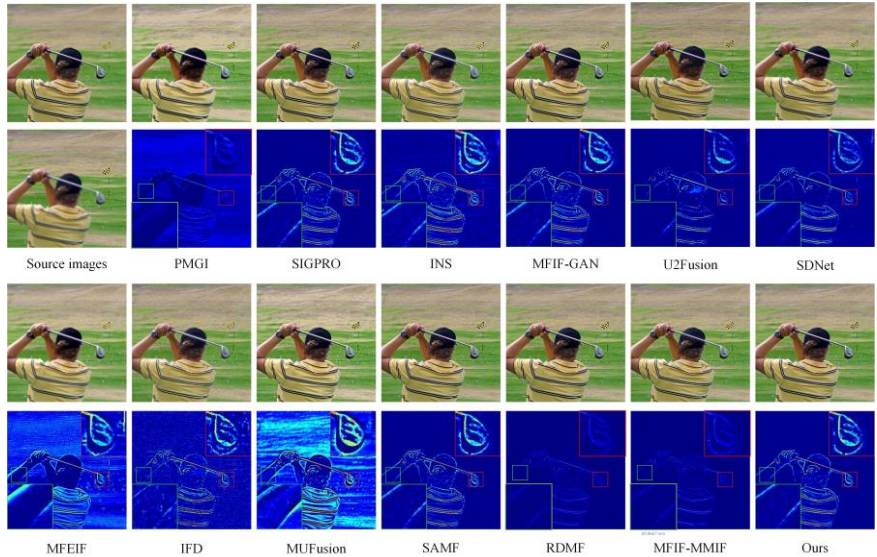

**Fig. 2** The different method in the test images "lytro-01"

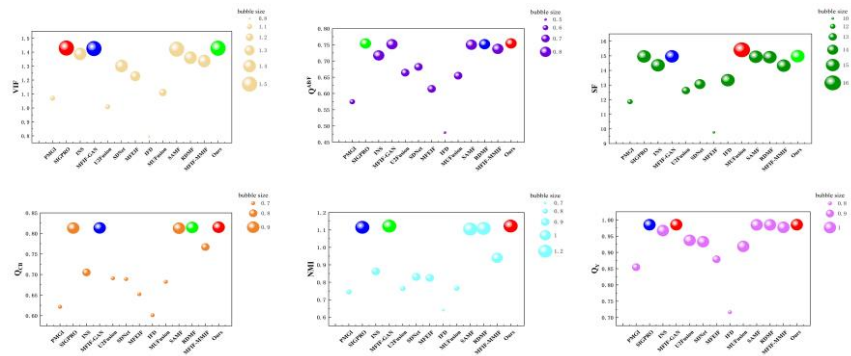

**Fig. 3** Quantitative analysis of the fusion method for the source image "lytro-01" (where the bubble size q represents the range of the metric within range $(q - 0.1)\sim q$, and specifically for the SF, within range $(q - 1)\sim q$.)

In comparison, the fusion results of the SIGPRO, MFIF-GAN, U2Fusion, SAMF, RDMF and MFIF-MMIF methods show improved edge representation, but some issues persist. Due to improper processing of edge information, some artifacts and burrs appear on the edge of the club in the red frames of the SIGPRO and SAMF pseudo-color images. In the U2Fusion ,RDMF and MFIF-MMIF pseudo-color images, misjudgment between the defocused area and the focus area results in the missing edges around the arm or the boundary is not obvious. Among all methods, the fused images produced by MFIF-GAN and the proposed method show superior visual quality, preserving smooth and clear boundaries. As depicted in Figure 3, our method ranks in the top second across six metrics—achieving the highest score in four—demonstrating its advantage in accurately preserving focus-defocus boundaries and reliably identifying focus regions compared to 12 state-of-the-art techniques.

### 3.2 Quantitative comparisons

Tables 1-4 show quantitative comparisons of the methods presented in this paper with 12 other SOTA methods in the four datasets, respectively. In the Lytro dataset, all metrics of the proposed method ranked in the top two, with the highest scores in VIF and QCB metrics. Even if the source image of the MFFW dataset has a boundary DSE, our method can still handle the boundary problem better and obtain clear and accurate boundaries. All indicators of the proposed method are in the top three. The images in the MFI-WHU dataset have complex boundaries, and the proposed method can effectively extract boundary information, so it always ranks in the top two in the indicators. In the GrayScale dataset, even if affected by grayscale images, the proposed method can better express images in the MFI-WHU dataset have complex boundaries, and the proposed method can effectively extract boundary information, so it always ranks in the top two in the indicators. In the GrayScale dataset, even if affected by grayscale images, the proposed method can better express boundaries. Except for the NMI indicator, all other indicators rank in the top three. And from the comprehensive score of the indicators, the proposed method ranks first on the above four datasets. Therefore, based on quantitative evaluation, we concluded that the overview method outperforms other SOTA methods.

**Table 1** Objective performance of different fusion methods on Lytro dataset

| Method | VIF | Q$^{ABF}$ | SF | Q$_{CB}$ | NMI | Q$_V$ | Score | Rank |
|---|---|---|---|---|---|---|---|---|
| PMGI | 0.9634 | 0.5058 | 12.1405 | 0.5873 | 0.8001 | 0.8098 | 12 | 13 |
| SIGPRO | 1.3633 | 0.7530 | 19.4040 | 0.8018 | 1.1261 | 0.9793 | 67 | 3 |
| INS | 1.1477 | 0.7140 | 18.6682 | 0.6690 | 0.8931 | 0.9510 | 39 | 7 |
| MFIF-GAN | 1.3630 | 0.7529 | 19.4271 | 0.8005 | 1.1313 | 0.9797 | 68 | 2 |
| U2Fusion | 1.0248 | 0.6337 | 15.3529 | 0.6462 | 0.7891 | 0.8855 | 18 | 12 |
| SDNet | 1.1241 | 0.6803 | 17.8238 | 0.6519 | 0.8520 | 0.9076 | 31 | 9 |
| MFEIF | 1.0655 | 0.5779 | 11.9865 | 0.6096 | 0.8549 | 0.8304 | 19 | 11 |
| IFD | 1.0341 | 0.6356 | 21.8814 | 0.5767 | 0.6816 | 0.6871 | 23 | 10 |
| MUFusion | 1.1504 | 0.6624 | 18.9482 | 0.6770 | 0.7983 | 0.9068 | 34 | 8 |
| SAMF | 1.3575 | 0.7511 | 19.3820 | 0.7951 | 1.1191 | 0.9770 | 54 | 5 |
| RDMF | 1.3679 | 0.7518 | 19.3444 | 0.8010 | 1.1221 | 0.9790 | 62 | 4 |
| MFIF-MMIF | 1.2576 | 0.7318 | 18.5671 | 0.7338 | 0.9329 | 0.9632 | 45 | 6 |
| Ours | 1.3644 | 0.7531 | 19.4306 | 0.8024 | 1.1292 | 0.9796 | 74 | 1 |

**Table 2** Objective performance of different fusion methods on MFFW dataset

| Method | VIF | Q$^{ABF}$ | SF | Q$_{CB}$ | NMI | Q$_V$ | Score | Rank |
|---|---|---|---|---|---|---|---|---|
| PMGI | 0.8142 | 0.4698 | 13.8816 | 0.5262 | 0.7267 | 0.7097 | 10 | 13 |
| SIGPRO | 0.9732 | 0.6360 | 22.6809 | 0.6867 | 0.7749 | 0.8540 | 68 | 2 |
| INS | 0.85942 | 0.6069 | 21.6675 | 0.5993 | 0.7559 | 0.8340 | 43 | 6 |
| MFIF-GAN | 0.9776 | 0.6306 | 22.4090 | 0.6884 | 0.7735 | 0.8512 | 65 | 3 |
| U2Fusion | 0.7953 | 0.5461 | 16.5798 | 0.5786 | 0.7187 | 0.7577 | 19 | 11 |
| SDNet | 0.8578 | 0.5765 | 21.7309 | 0.5776 | 0.7478 | 0.7883 | 34 | 9 |
| MFEIF | 0.9285 | 0.5540 | 13.5768 | 0.5753 | 0.8132 | 0.8016 | 35 | 8 |
| IFD | 0.6902 | 0.4671 | 19.6419 | 0.5316 | 0.6104 | 0.6067 | 12 | 12 |
| MUFusion | 0.8537 | 0.5596 | 20.1091 | 0.6087 | 0.7199 | 0.7887 | 30 | 10 |
| SAMF | 0.9512 | 0.6292 | 22.6846 | 0.6693 | 0.7368 | 0.8125 | 51 | 5 |
| RDMF | 0.9837 | 0.6296 | 22.145 | 0.6908 | 0.7665 | 0.8499 | 64 | 4 |

| Method | VIF | Q^ABF | SF | Q_CB | NMI | Q_Y | Score | Rank |
|---|---|---|---|---|---|---|---|---|
| MFIF-MMIF | 0.9064 | 0.6219 | 21.3722 | 0.6375 | 0.7473 | 0.8327 | 43 | 6 |
| Ours | 0.9752 | 0.6361 | 22.8127 | 0.6888 | 0.7742 | 0.8528 | 72 | 1 |

Table 3 Objective performance of different fusion methods on MFI-WHU dataset

| Method | VIF | Q^ABF | SF | Q_CB | NMI | Q_Y | Score | Rank |
|---|---|---|---|---|---|---|---|---|
| PMGI | 0.9733 | 0.5080 | 17.0356 | 0.6163 | 0.7552 | 0.7959 | 15 | 12 |
| SIGPRO | 1.3808 | 0.7280 | 26.7604 | 0.8222 | 1.1845 | 0.9835 | 63 | 3 |
| INS | 1.3274 | 0.7230 | 26.6967 | 0.7786 | 1.0381 | 0.9739 | 49 | 6 |
| MFIF-GAN | 1.3786 | 0.7327 | 26.8503 | 0.8227 | 1.1841 | 0.9847 | 67 | 2 |
| U2Fusion | 0.9622 | 0.5729 | 18.9410 | 0.6240 | 0.7079 | 0.8652 | 18 | 11 |
| SDNet | 1.2428 | 0.6856 | 26.2088 | 0.7196 | 0.8816 | 0.9448 | 37 | 8 |
| MFEIF | 1.1031 | 0.5662 | 15.6286 | 0.6846 | 0.8501 | 0.848 | 22 | 10 |
| IFD | 0.7830 | 0.4760 | 24.5354 | 0.6045 | 0.6179 | 0.6725 | 10 | 13 |
| MUFusion | 1.0656 | 0.5996 | 22.5857 | 0.6473 | 0.7287 | 0.8737 | 25 | 9 |
| SAMF | 1.3775 | 0.7263 | 26.6674 | 0.8203 | 1.1883 | 0.9849 | 62 | 4 |
| RDMF | 1.3837 | 0.7289 | 26.5912 | 0.8236 | 1.1743 | 0.9828 | 62 | 4 |
| MFIF-MMIF | 1.3174 | 0.7173 | 25.8464 | 0.7915 | 1.0381 | 0.9706 | 42 | 7 |
| Ours | 1.3848 | 0.7330 | 26.7834 | 0.8253 | 1.1850 | 0.9848 | 74 | 1 |

**Table 4** Objective performance of different fusion methods on GrayScale dataset

| Method | VIF | Q^ABF | SF | Q_CB | NMI | Q_Y | Score | Rank |
|---|---|---|---|---|---|---|---|---|
| PMGI | 0.9032 | 0.5334 | 16.6665 | 0.5774 | 0.7498 | 0.7535 | 12 | 12 |
| SIGPRO | 1.0657 | 0.5195 | 17.3427 | 0.6616 | 0.9692 | 0.8879 | 47 | 6 |
| INS | 1.0004 | 0.6306 | 22.2542 | 0.6521 | 0.8183 | 0.8679 | 40 | 8 |
| MFIF-GAN | 1.0997 | 0.6424 | 22.5395 | 0.7077 | 0.8809 | 0.8814 | 68 | 2 |
| U2Fusion | 0.8767 | 0.5559 | 16.6350 | 0.6130 | 0.7238 | 0.7810 | 12 | 12 |
| SDNet | 0.9535 | 0.5792 | 23.3981 | 0.6270 | 0.7657 | 0.8232 | 36 | 9 |
| MFEIF | 0.9673 | 0.5556 | 13.9899 | 0.6153 | 0.8215 | 0.8193 | 22 | 11 |
| IFD | 1.0341 | 0.6356 | 21.8814 | 0.6742 | 0.8237 | 0.8693 | 48 | 5 |
| MUFusion | 0.9965 | 0.5777 | 20.3517 | 0.6212 | 0.7289 | 0.8014 | 24 | 10 |
| SAMF | 1.077 | 0.6382 | 22.4320 | 0.7022 | 0.8649 | 0.8764 | 58 | 4 |
| RDMF | 1.1011 | 0.6413 | 22.4109 | 0.7061 | 0.8816 | 0.8853 | 67 | 3 |
| MFIF-MMIF | 1.0149 | 0.6301 | 21.4024 | 0.6675 | 0.7867 | 0.8706 | 41 | 7 |
| Ours | 1.1010 | 0.6446 | 22.5101 | 0.7167 | 0.8804 | 0.8870 | 71 | 1 |

Based on the scores and rankings in Tables 1-4, our method achieved the highest overall scores in all four public datasets, and performed particularly well in the Lytro and MFI-WHU datasets. To sum up, our method exhibits the highest fusion performance compared to the 12 state-of-the-art methods.

### 3.3 Parameter analysis

This section mainly assesses the influence of the parameter $th$ in the adaptive threshold "bwareaopen" filling filter in Equation (9) and the parameter $k$ in Equation (6) on the fusion results. To assess the influence of the parameter th on fusion performance, we fixed k at 0.5 and then examined the effect of varying th. The fusion outcomes were quantitatively evaluated using the $Q_Y$, VIF, and $Q_{CB}$ metrics. The average scores of different th values in the data set Lytro are shown in Figure 5. When $th = 0.02$, a significant improvement is observed in the overall fusion performance, so based on comprehensive analysis, we set the parameter $th = 0.02$. After setting parameter $th = 0.02$, we analyze the impact of parameter $k$ on fusion performance. To more intuitively observe the impact of parameters on the fusion results, we use the input image pair "lytro-01" in the Lytro data set as the experimental image to assess the impact of different parameters k on the final decision map, as shown in Figure 6. In Figure 6, it can be found that when $k$ is greater than 0, the edge of the final decision is more accurate, so introducing parameter $k$ is effective. When $k = 0.5$, some small structures on the edge almost completely disappear and edge information is retained. Comprehensive analysis of objective evaluation and subjective vision shows that when $k = 0.5$ it has good fusion performance, so the parameter $k$ is set 0.5. Therefore, we set the two key parameters $th$ and $k$ to 0.02 and 0.5 respectively.

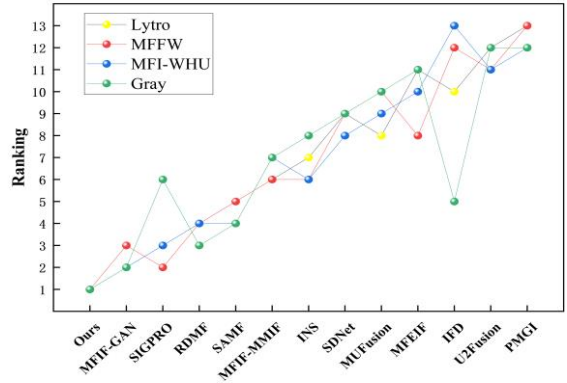

**Fig. 4** Ranking of the 12 MFIF methods and the proposed method on the four datasets and the overall ranking (The proposed method ranking first.)

### 3.4 ablation experiments

In the ablation studies,we conducted on the Lytro dataset to evaluate the impact of four key components: the edge enhancement strategy of the initial fused image, the adaptive threshold "bwareaopen" filling filter, guided filtering, and consistency verification. During testing, all settings were kept consistent across modules, except for the specific ablation module under evaluation. The experiments reveal the crucial importance of the proposed salient information enhancement and the employed post-processing strategies.

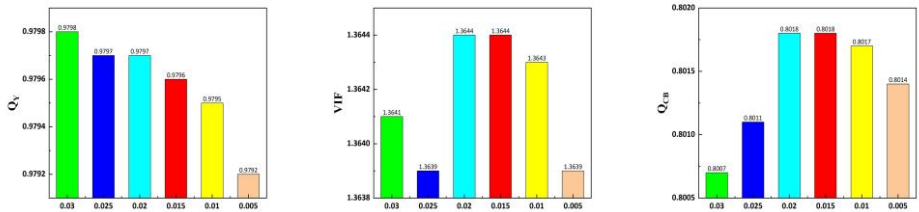

**Fig. 5** Effect of parameter th on the fusion performance of the proposed method on the Lytro dataset. The index of the horizontal axis indicates the value of the parameter $th$.

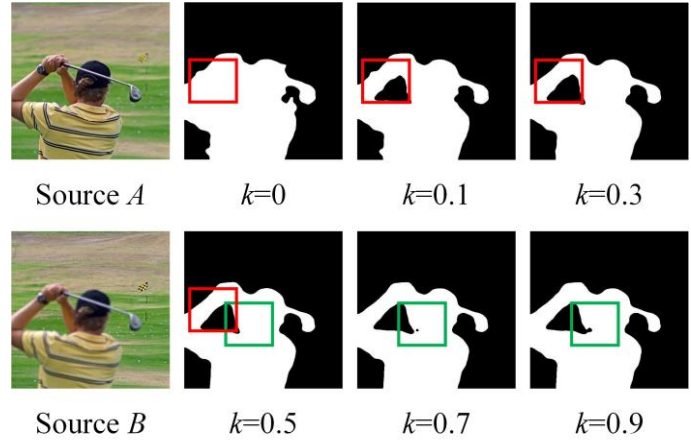

**Fig. 6**. Obtain the final decision maps with different parameters $k$ in source images A, B.

First, the saliency map of one input image is used in Eq. (5) instead of the initial fused image to enhance the saliency map of another input image. Second, we sequentially remove the adaptive thresholding "bwareaopen" filler filter, bootstrap filtering, and consistency validation, and verify its effectiveness. Table 5 presents the objective evaluation results of ablation experiments for each model on the Lytro dataset, demonstrating that the putted forward method outperforms all others across all metrics. To better illustrate the impact of each module on fusion performance, we use the input image pair "lytro-03" from the Lytro dataset to analyze the effect of different modules on the final decision map, as shown in Figure 7. In Figure 7, boundary issues are visible in decision maps(b)-(e). In contrast, decision map (a), generated by the proposed method, avoids these issues and achieves smooth, clear boundaries. In summary, any substitution or omission of parts results in block effects, blurring, and reduced overall fusion performance in the final decision map. These findings confirm that each component in the putted forward method is effective and significantly enhances fusion quality and performance.

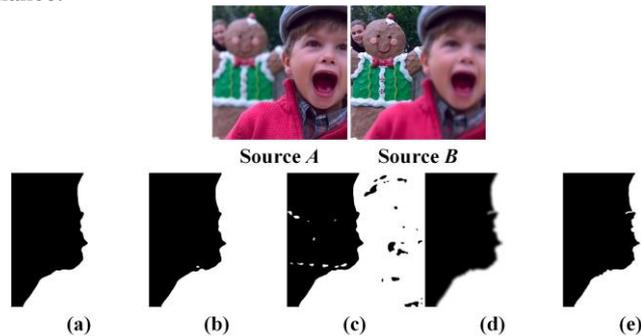

**Fig. 7** The final decision maps obtained by ablating the modules in the proposed method. (a) the proposed method, (b) without significant information enhancement, (c) without the "bwareaopen" filter , (d) without guided filtering, (e) without consistency verification.

**Table 5.** Mean metric results of ablation experiments on the Lytro dataset. "\" indicates that the component was not used, and an "√" indicates that the component is used.

| Index | | 1 | 2 | 3 | 4 | 5(Ours) |
|---|---|---|---|---|---|---|
| initial fusion image | | \ | √ | √ | √ | √ |
| "bwareaopen" filters | | √ | √ | √ | √ | √ |
| guided filters | | √ | √ | √ | √ | √ |
| consistency verification | | √ | √ | √ | \ | √ |
| metric | VIF | 1.3541 | 1.3628 | **1.3648** | 1.3620 | *1.3644* |
| | Q^{AB/F} | 0.7509 | 0.7530 | **0.7531** | 0.7520 | **0.7531** |
| | SF | 19.3333 | **19.4366** | 19.4261 | 19.2639 | *19.4306* |
| | Q_{CB} | 0.7988 | 0.7998 | **0.8012** | 0.7995 | **0.8018** |
| | NMI | **1.1299** | 1.1290 | **1.1299** | 1.1125 | **1.1303** |
| | Q_Y | **0.9800** | 0.9785 | 0.9794 | 0.9789 | *0.9797* |

## 4    Conclusions

This paper proposed in a MFIF method based on gradient detection and saliency enhancement. To obtain complete and clear focused edge information, we design a strategy based on salient edge enhancement. And we obtain an initial decision map with more accurate edges through through the GICI module. We used six commonly used metrics to conduct qualitative and quantitative analysis in the four datasets, and compared them with 12 SOTA methods. The consistency of the experimental results shows that our method has obvious advantages in both subjective and objective evaluations, and the fusion result can better retain edge details and effectively solve the problem of boundary troubles.